\documentclass[letterpaper, 10 pt, conference]{ieeeconf}  

\IEEEoverridecommandlockouts                              

\usepackage{amssymb}  

\usepackage{graphicx}
\usepackage{algorithm}
\usepackage{algcompatible}
\usepackage{amsfonts}
\usepackage{amsmath}
\usepackage{hhline}

\usepackage{subcaption}

\usepackage{bbm}
\usepackage{dsfont}

\usepackage{makecell}

\newcommand{\vect}[1]{{\boldsymbol{#1}}}

\newcommand{\E}{\mathbb{E} }
\newcommand{\R}{\mathbb{R} }

\newcommand{\KL}{D_{\textrm{KL}}}

\newcommand{\new}{\textrm{new}}

\newcommand{\demo}{\textrm{demo}}
\newcommand{\joint}{\textrm{joint}}
\newcommand{\sample}{\textrm{sample}}

\title{\LARGE \bf
Goal-Conditioned  Variational Autoencoder Trajectory Primitives \\ with Continuous and Discrete Latent Codes
}

\author{Takayuki Osa$^{1,2}$ and Shuehi Ikemoto$^{1}$
\thanks{$^{1}$T. O. and S. I. are with Department of Human Intelligence,
        Kyushu Institute of Technology, 808-0135, Fukuoka, Japan.
        {\tt\small $\{$ osa, ikemoto $\}$@brain.kyutech.ac.jp}}%
\thanks{$^{2}$T. O. is also with RIKEN center for Advanced Intelligence Project, Tokyo, Japan.}%
}

\begin{document}

\maketitle
\thispagestyle{empty}
\pagestyle{empty}

\begin{abstract}
Imitation learning is an intuitive approach for teaching motion to robotic systems.
Although previous studies have proposed various methods to model demonstrated movement primitives,
one of the limitations of existing methods is that the shape of the trajectories are encoded in high dimensional space.
The high dimensionality of the trajectory representation can be a bottleneck in the subsequent process such as planning a sequence of primitive motions.
We address this problem by learning the latent space of the robot trajectory.
If the latent variable of the trajectories can be learned, it can be used to tune the trajectory in an intuitive manner even when the user is not an expert.
We propose a framework for modeling demonstrated trajectories with a neural network that learns the low-dimensional latent space.
Our neural network structure is built on the variational autoencoder~(VAE) with discrete and continuous latent variables.
We extend the structure of the existing VAE to obtain the decoder that is conditioned on the goal position of the trajectory for generalization to different goal positions.
Although the inference performed by VAE is not accurate, the positioning error at the generalized goal position can be reduced to less than 1~mm by incorporating the projection onto the solution space.
To cope with requirement of the massive training data, we use a trajectory augmentation technique inspired by the data augmentation commonly used in the computer vision community.
In the proposed framework, the latent variables that encodes the multiple types of trajectories are learned in an unsupervised manner, although existing methods usually require label information to model diverse behaviors.
The learned decoder can be used as a motion planner in which the user can specify the goal position and the trajectory types by setting the latent variables. 
The experimental results show that our neural network can be trained using a limited number of demonstrated trajectories and 
that the interpretable latent representations can be learned.

\end{abstract}

\section{INTRODUCTION}
\begin{figure}[]
	\centering
	\includegraphics[width=\columnwidth]{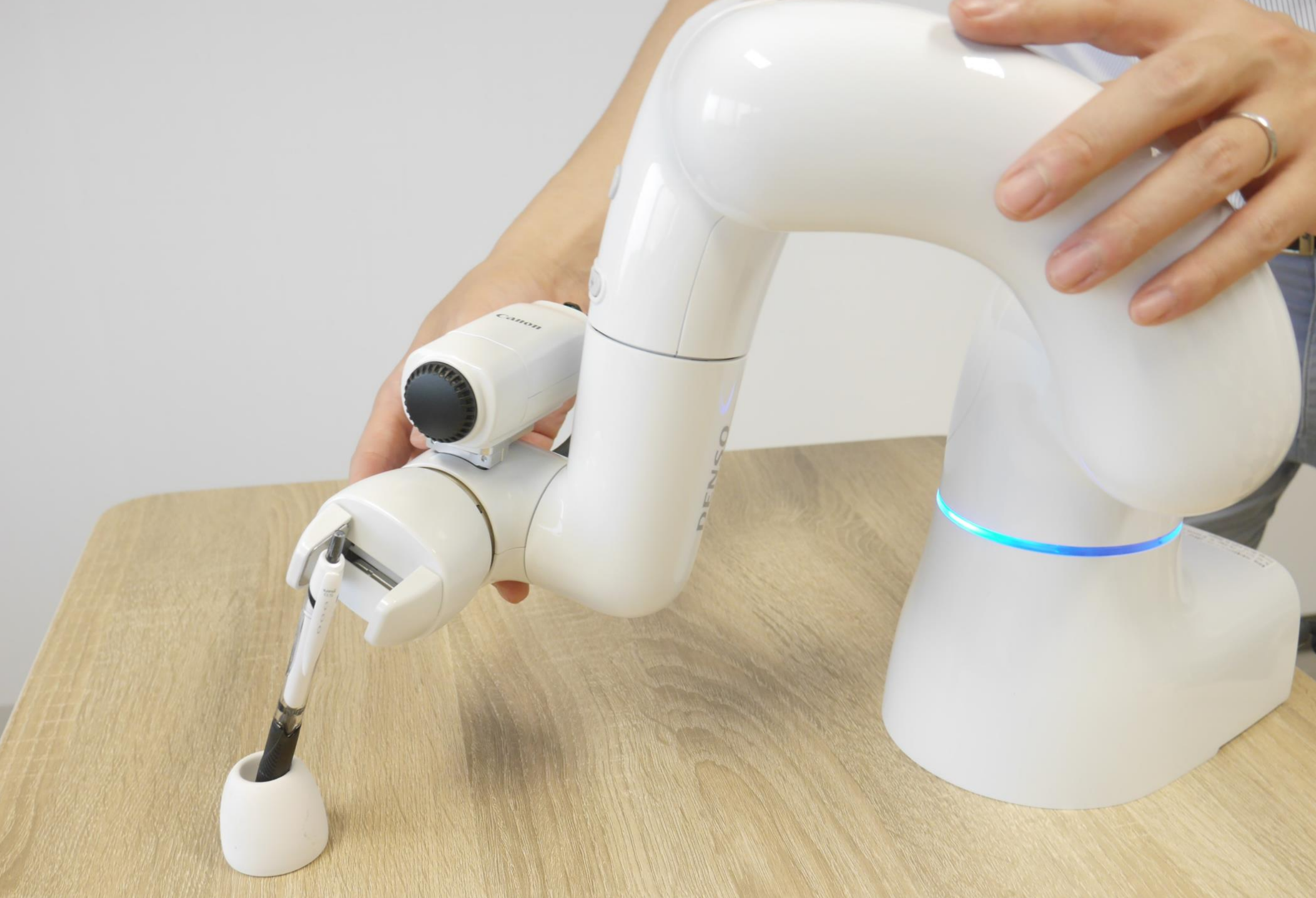}
	\caption{Modeling the demonstrated trajectory through kinesthetic teaching is essential in imitation learning.}
	\label{fig:intro}
\end{figure}

Imitation learning~(IL)~\cite{Osa18b} is an approach that can achieve such intuitive motion teaching. In IL, a user demonstrates how a task is performed, e.g. through kinesthetic teaching (Fig.~\ref{fig:intro}), and the system learns how to generalize demonstrated trajectories to different conditions.
Previous studies have proposed various ways to model demonstrated trajectories such as Dynamic Movement Primitive~(DMP)~\cite{Ijspeert02} and Probabilistic Movement Primitive~(ProMP)~\cite{Paraschos13} and Kernelized Movement Primitive~(KMP)~\cite{Huang19}. 
Although these methods can cope with complex generalization of the demonstrated trajectories, one of the limitations of existing methods is that the shape of the trajectories are encoded in high dimensional space.
A trajectory of a robotic manipulator is often high-dimensional. 
For example, if we represent a trajectory of a manipulator with 7 DoFs as 100 way points, a trajectory will be represented as a vector with 700 dimensions.
Even if we project a trajectory onto a weight space as in existing methods such as DMP~\cite{Ijspeert02}, the trajectory will be encoded as a weight vector with hundreds of dimensions.
The high dimensionality of the trajectory representation can be a bottleneck in the subsequent process such as planning a sequence of primitive motions.
We address this problem by learning the latent space of the robot trajectory.
If the low-dimensional latent variable of the trajectories can be learned, it can be used to tune the trajectory when planning and optimizing a sequence of primitive trajectories.
In addition, by encoding the trajectory type into the latent variable, we can model multiple trajectory types with a single model, although existing frameworks usually require separate multiple models to represent multiple behaviors.

Deep learning, which has contributed to recent advances in machine learning, is a powerful tool for coping with high-dimensional data. Moreover, it can be used to learn latent representations that capture meaningful information as low-dimensional data~\cite{Kingma14,Goodfellow14}. 
However, neural networks are notorious in that they require thousands of samples for training.
In the context of imitation learning, it is costly to collect data of human demonstrations. 
Therefore, the number of the available expert trajectories is often limited.
To leverage the power of neural networks in imitation learning, it is essential to address the requirement of massive training data.

In this study, we present a framework for modeling multiple types of demonstrated trajectories with a single neural network by learning latent representations.
The proposed framework does not require information on the motion type of the demonstrated trajectories, and the model is trained in an unsupervised manner.
To address the issue of the size of the training data, we propose a trajectory augmentation trick that is inspired by the data augmentation commonly used in the computer vision community.
We also extend the structure of the joint VAE proposed in~\cite{Dupont18} to obtain a decoder that is conditioned on the goal position of the trajectory for generalization to different goal positions.
In this framework, different types of trajectories are encoded in the discrete latent variable and the continuous variable interpolates the input trajectories. 
The learned decoder can be used as a motion planner in which the user can specify the goal position and the trajectory types by setting the latent variables. 
The experimental results show that the proposed neural network can be trained with a limited number of demonstrated trajectories and learns interpretable low-dimensional representations.
We think that our framework will lead to the development of an imitation-learning-based robotic system in which a user can select and modify the shape of a planned trajectory by changing low-dimensional latent variables.

\section{RELATED WORK}
%
Imitation learning is a class of methods for learning the behavior demonstrated by experts~\cite{Osa18b}, 
and it is considered an intuitive approach that facilitates the teaching of desired motions to a robotic system.
Motion planning with imitation learning has been exploited in practical applications such as robotic surgery~\cite{Osa18}. 
Previous studies on movement primitives have proposed various frameworks for modeling demonstrated trajectories, e.g. DMP~\cite{Ijspeert02}, ProMP~\cite{Paraschos13}, and KMP~\cite{Huang19}.
These methods enable the generalization of trajectories to different goal positions while maintaining the topological features of the demonstrated trajectories.
However, these methods do not learn lower-dimensional latent representations of demonstrated trajectories, which would enable intuitive user-system interaction.
Studies on deep generative models~\cite{Kingma14,Goodfellow14,Dupont18} show that it is possible to learn intuitive latent representations of data.
The original study of VAE shows that VAE can learn the latent dimension which corresponds to the facial expression and that the face image can be continuously changed from a ``sad'' face to an ``angry'' face by changing the value of latent variable~\cite{Kingma14}.  
In the context of imitation learning, it will be possible to generate a new ``middle'' behavior by learning latent representations of demonstrated trajectories.

The latent space of robotic trajectories has been leveraged in the context of imitation learning.
Studies such as \cite{Shon05,Grimes08} employed Gaussian Process Latent Variable Model~\cite{Lawrence05} and Principal Component Analysis to learn a low-dimensional latent space to achieve efficient learning.
However, it is not trivial to learn both continuous and discrete latent variables using these methods.
In contrast, the autoencoder used in this study learns both discrete and continuous latent variables.

Recent advances in reinforcement learning (RL) has made it possible to apply deep learning methods to robotic manipulation.
The study in~\cite{Levine16b} trained a neural network to generate a motor torque command from image inputs.
Studies such as \cite{Haarnoja18} proposed algorithms to learn neural network policies using deep RL for real robots.
In these recent studies, the use of neural networks is often focused on Markov Decision Process~(MDP) settings, where the agent performs learning to generate a control input from a given state under stochastic dynamics.
The study in~\cite{Merel19} is related to our work in that a neural network learns to imitate demonstrated behaviors through learning the latent variable for continuous control tasks.  
However, the use of neural networks for planning problems is still unexplored. 
In planning problems, it is usually unnecessary to consider a stochastic state transition as in the standard RL setting~\cite{Levine18}.
Instead, a desired trajectory is planned for a given deterministic state transition, which is actually valid assumption in practice.
Many industrial robots have solid position/velocity controllers, and there is no need to learn low-level control. 
It is natural to consider that the system is fully actuated in these cases.
Therefore, we think that control of industrial robots requires an algorithm that addresses the planning problem rather than the continuous control problem.

A few recent studies have addressed the application of neural networks to manipulation planning.
The study in~\cite{Arnold19} proposed a neural network architecture that predicts the deformation of objects after a simple manipulation. 
To train the neural network, hundreds of manipulation trajectories were recorded. 
The necessity of massive training data has been an issue of applying deep learning to motion planning.
In our study, we propose a technique for generating synthetic trajectories to address this issue.

\section{AUTOENCODER TRAJECTORY PRIMITIVES}
We present the proposed autoencoder to model the demonstrated trajectories, which we refers to as a \textit{autoencoder trajectory primitive}~(ATP).
The network architecture and the objective function are described.
Then the process of creating a dataset with a sufficient number of trajectories from a limited number of demonstrated trajectories is described.

\subsection{Objective Function and Architecture}
We denote by $\vect{q} \in \R^d$ a configuration of a robot manipulator, where $d$ is a number of joints.
A trajectory given by a sequence of configurations is denoted by
$\vect{\xi} = [\vect{q}_0, \ldots, \vect{q}_T]$ where $T$ is the number of time steps.  
We assume that a dataset of trajectories $\mathcal{D} = \{ \vect{\xi}_i \}^N_{i=1}$ is given.
It is also assumed that $N$ is a sufficiently large to train a neural network.
We describe the process of constructing such a dataset from a limited number of demonstrated trajectories in the next section.
In this study, we aim to learn the latent space of a given dataset.
The continuous and discrete latent variables are denoted by $\vect{z}$ and $\vect{c}$, respectively.
We consider the variational autoencoder framework in which the the posterior/encoder $ q_{\vect{\phi}}(\vect{z}, \vect{c} | \vect{\xi})$ is represented by a neural network parameterized by a vector $\vect{\phi}$, and the likelihood/decoder $p_{\vect{\theta}} (\vect{\xi}|\vect{z}, \vect{c})$ is represented by a neural network parameterized by a vector $\vect{\theta}$.
For learning latent variables, $\beta$-VAE model~\cite{Higgins16} is often employed in previous studies. 
When learning the continuous and discrete variable $\vect{z}$ and $\vect{c}$, the objective function of the $\beta$-VAE model~\cite{Higgins16} is given by
\begin{align}
\mathcal{J}_{\beta} = \E_{\vect{\xi} \sim \mathcal{D}} \left[
\mathcal{L}_{\beta}(\vect{\theta}, \vect{\phi})
\right],
\end{align}
where $\mathcal{L}_{\beta}(\vect{\theta}, \vect{\phi})$ is given by
\begin{align}
\mathcal{L}_{\beta}(\vect{\theta}, \vect{\phi}) = \E_{q_{\vect{\phi}}(\vect{z}, \vect{c} | \vect{\xi})} 
[ \log p_{\vect{\theta}} (\vect{\xi}|\vect{z}, \vect{c}) ] \nonumber \\
- \beta \KL( q_{\vect{\phi}}(\vect{z}, \vect{c} | \vect{\xi}) || p(\vect{z}, \vect{c}) ), 
\end{align}
where $q_{\vect{\phi}}(\vect{z}, \vect{c} | \vect{\xi})$ is the joint posterior, $p(\vect{z}, \vect{c})$ is the prior, $ p_{\vect{\theta}} (\vect{\xi}|\vect{z}, \vect{c})$ is the likelihood, and 
$\KL( q_{\vect{\phi}}(\vect{z}, \vect{c} | \vect{\xi}) || p(\vect{z}, \vect{c}) )$ is the KL divergence between $q_{\vect{\phi}}(\vect{z}, \vect{c} | \vect{\xi})$ and $p(\vect{z}, \vect{c}) $.
However, the latent variable learned by $\beta$-VAE is often non-intuitive. 
For learning disentangled continuous and discrete representations, Dupont\cite{Dupont18} proposed the objective 
given as
\begin{align}
\mathcal{J}_{\joint} = \E_{\vect{\xi} \sim \mathcal{D}} \left[
\mathcal{L}_{\joint}(\vect{\theta}, \vect{\phi})
\right],
\end{align}
where $\mathcal{L}_{\joint}(\vect{\theta}, \vect{\phi})$ is given by
\begin{align}
\mathcal{L}_{\joint}(\vect{\theta}, \vect{\phi}) = \E_{q_{\vect{\phi}}(\vect{z}, \vect{c} | \vect{\xi})} 
[ \log p_{\vect{\theta}} (\vect{\xi}|\vect{z}, \vect{c}) ] \nonumber \\
- \gamma \left| \KL\big( q_{\vect{\phi}}(\vect{z} | \vect{\xi}) || p(\vect{z}) \big) - C_{\vect{z}} \right| \nonumber  \\
- \gamma \left| \KL\big( q_{\vect{\phi}}( \vect{c} | \vect{\xi}) || p(\vect{c}) \big) - C_{\vect{c}} \right|,
\end{align}
where $C_{\vect{z}}$ and $C_{\vect{c}}$ represent the information capacity, and $\gamma$ is a coefficient.
In our framework, we can compute the end-effector position at the goal configuration $\vect{x}_g$.
This supervised information $\vect{x}_g$ is incorporated into the proposed neural network to obtain the decoder $p_{\vect{\theta}} (\vect{\xi}|\vect{z}, \vect{c}, \vect{x}_g)$ conditioned on $\vect{x}_g$. 
This supervised auxiliary code enables the generalization of the trajectories to different goal positions.
We employ the objective function given by
\begin{align}
\mathcal{J} = \E_{\vect{\xi} \sim \mathcal{D}} \left[
\mathcal{L}(\vect{\theta}, \vect{\phi}) - \log  q_{\vect{\phi}}( \vect{x}_g | \vect{\xi})
\right]
\label{eq:loss}
\end{align}
where 
\begin{align}
\mathcal{L}(\vect{\theta}, \vect{\phi}) & =  \E_{q_{\vect{\phi}}(\vect{z}, \vect{c}, \vect{x}^g | \vect{\xi})} [ \log p_{\vect{\theta}} (\vect{\xi}|\vect{z}, \vect{c}, \vect{x}^g) ]   \nonumber \\
& \ - \gamma \left| \KL\big( q_{\vect{\phi}}(\vect{z} | \vect{\xi}) || p(\vect{z}) \big) - C_{\vect{z}} \right|  \nonumber \\
& \ - \gamma \left| \KL\big( q_{\vect{\phi}}( \vect{c} | \vect{\xi}) || p(\vect{c}) \big) - C_{\vect{c}} \right|.
\end{align}
It can be seen that our objective function is similar to that of the M2 model proposed by Kingma et al.~\cite{Kingma14b}.
The second term in \eqref{eq:loss} can also be viewed as a regularization term.
We use the reparametrization trick in~\cite{Kingma14} for the continuous latent variable $\vect{z}$.
To learn the discrete latent variable $\vect{c}$, we employ the reparametrization with the Gumbel Max trick as in~\cite{Maddison17}.

\begin{figure}[]
	\centering
	\includegraphics[width=\columnwidth]{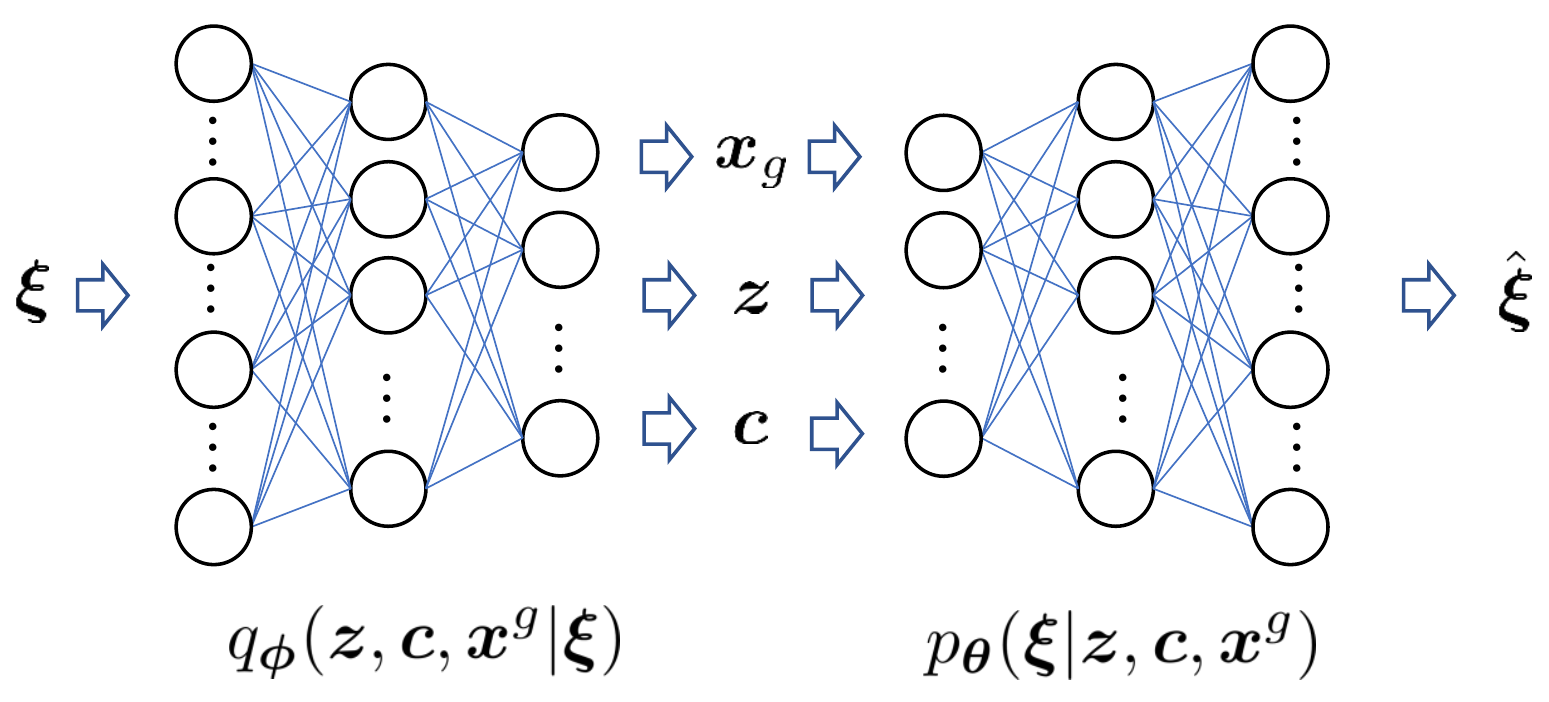}
	\caption{Network architecture of the proposed autoencoder trajectory primitive.
		$\vect{z}$ is the continuous latent code and $\vect{c}$ is the discrete latent code.}
	\label{fig:architecture}
\end{figure}

The network architecture is shown in Fig.~\ref{fig:architecture}.
The decoder obtained in our framework can be used as a user-interface for motion planning, in which the goal position can be specified from a given task and the user can tune the latent code to obtain a preferable shape of the trajectory. 
We employ fully-connected neural networks with two hidden layers for both the encoder and decoder.
The activation function was ReLU in our implementation.
In our implementation, input and reconstructed trajectories are represented in configuration space.
Therefore, inputs and outputs of the neural network are bounded between $-\pi$ to $\pi$.
We think this property is suitable for training a neural network; if the value of inputs and outputs are not bounded,
it would be necessary to use the batch normalization technique.

\subsection{Trajectory Augmentation}
\label{sec:augmentation}

When demonstrated trajectories $\{\vect{\xi}^{\demo}_i\}^M_{i=1}$ are given, we sample trajectories around the demonstrated trajectories to create a dataset $\mathcal{D}=\{\vect{\xi}_i\}^N_{i=1}$ with a sufficient number of trajectories to train the neural network. 
This approach is inspired by the data augmentation commonly used in the computer vision community.

To obtain various goal configurations, we sample perturbation at the goal configuration by following a Gaussian distribution
$\Delta \vect{q}_{T} \sim \mathcal{N} (\vect{0}, \Sigma_g)$
where $\Sigma_g$ is a covariance matrix, which is diagonal in our implementation.
The sampled perturbation is smoothly propagated to the whole trajectory as
\begin{align}
\Delta \vect{\xi}_g = M^{-1} [0, \ldots, 0, \Delta \vect{q}_{T}]^{\top}
\label{eq:goal_dist}
\end{align}
where $M^{\dagger}$ is the Moore-Penrose pseudo-inverse of the matrix $M$ defined by
\begin{align}
M  = 
\left[
\begin{array}{cccccc}
0 & 0  & 0 & \ldots & & 0 \\
0 &   2 & -1 &  & &\vdots\\ 
0  & -1 & 2 & -1  &  \ddots &\\
0 & 0 & -1 &  \ddots & 0 & 0\\
0 & 0 & & 2  & -1 & 0\\
\vdots& \vdots  &  \ddots  &-1 & 2  &- 1 \\
0 & 0 &\ldots & 0 & -1 & 2
\end{array}
\right]. 
\label{eq:metric}
\end{align}
The matrix $M$ is used in the trajectory update in CHOMP~\cite{Zucker13} and is used to project the trajectory onto the constraint trajectory space as we describe later.
$M$ plays a role in smoothly propagating the difference of the trajectory to the whole trajectory.

We also sample perturbation of the whole trajectory by following the distribution given as
\begin{align}
\beta_{\textrm{traj}}(\vect{\xi}) = \sum^{M}_{m=1}U(m)\mathcal{N}(\vect{\xi}^{\demo}_m, aB^{\dagger}),
\label{eq:exploration_stomp}
\end{align}	
where $a$ is a constant, $U(m)$ is the uniform distribution, and $B^{\dagger}$ is the Moore-Penrose pseudo-inverse of the matrix $M^{\top}M$.
In prior work~\cite{Kalakrishnan11}, the use of this covariance matrix for sampling trajectories was proposed, and it was empirically shown that using this covariance matrix leads to a smooth perturbation of the entire trajectory without changing the start and goal configurations.
Combining these techniques, we sample trajectories by following
\begin{align}
\vect{\xi}_{\sample} = \Delta \vect{\xi}_g + \vect{\xi}_{\beta} 
\label{eq:expl}
\end{align}
where $\Delta \vect{\xi}_g$ is obtained from \eqref{eq:goal_dist}, and $\vect{\xi}_{\beta}$ is obtained from $\beta_{\textrm{traj}}(\vect{\xi})$ given by \eqref{eq:exploration_stomp}.

\begin{figure}[]
	\centering
	\begin{subfigure}[t]{0.465\columnwidth}
		\includegraphics[width=\textwidth]{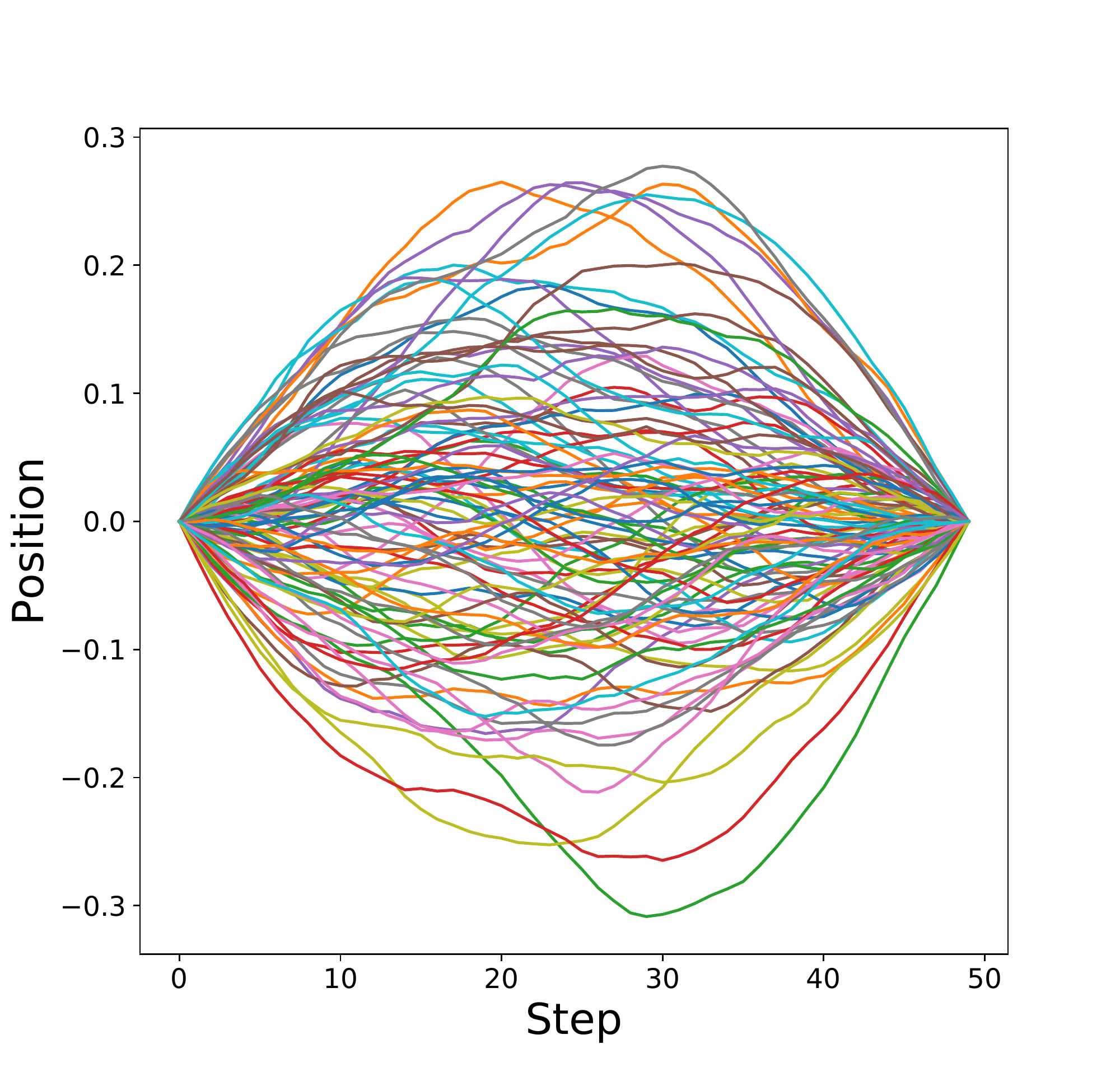}
		\caption{Trajectories obtained from  $\beta_{\textrm{traj}}(\vect{\xi})$ given by \eqref{eq:exploration_stomp}. }
	\end{subfigure}
	\begin{subfigure}[t]{0.465\columnwidth}
		\includegraphics[width=\textwidth]{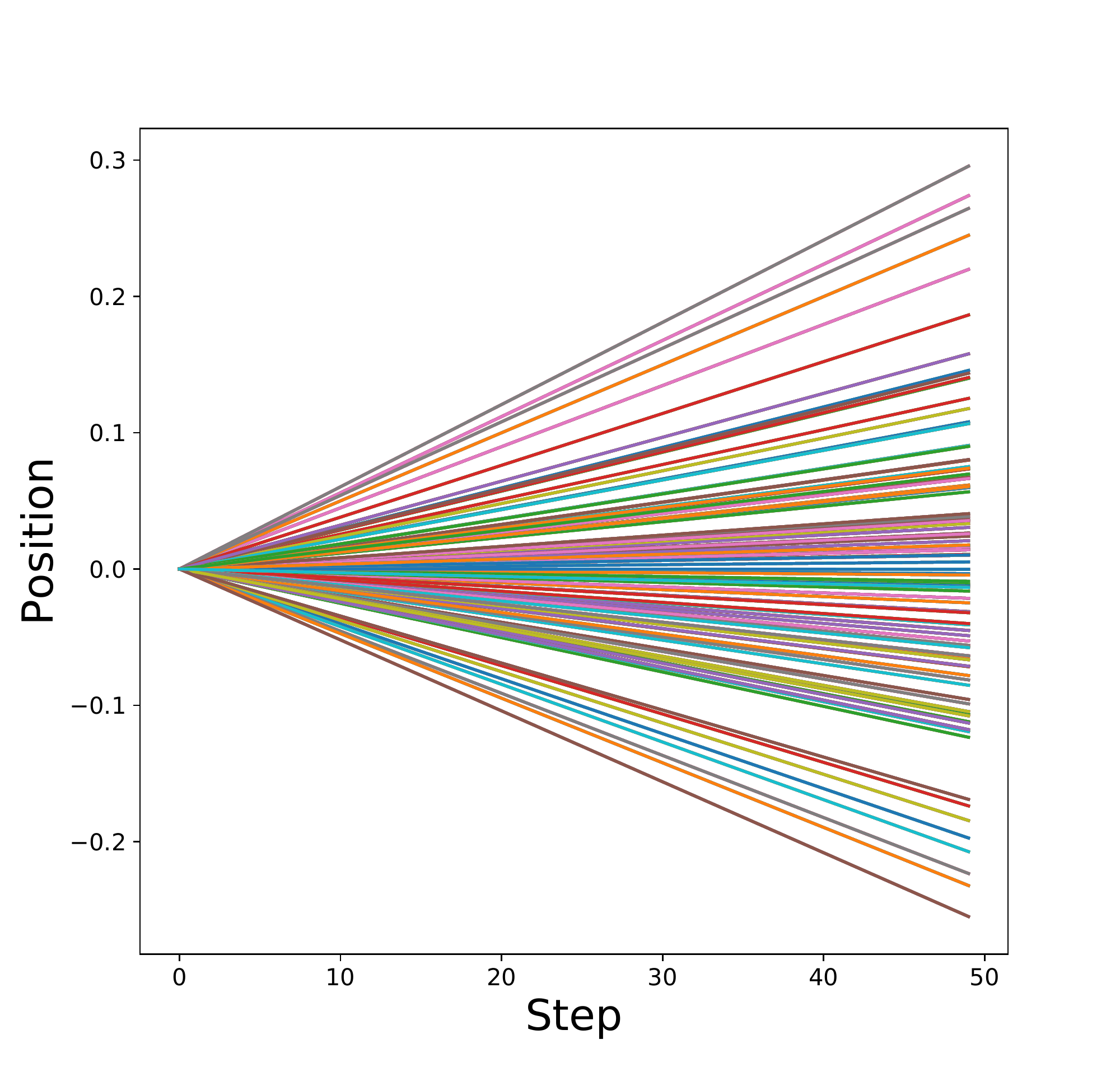}
		\caption{Trajectories obtained from \eqref{eq:goal_dist}.}
	\end{subfigure}
	\caption{Visualization of trajectories sampled for trajectory augmentation. }
	\label{fig:exploration}
\end{figure}

\subsection{Projection onto the Constraint Space}
Although our decoder generates a trajectory from a given goal position, the generated trajectory does not precisely satisfy the given goal position.
However, it is difficult to explicitly incorporate the constraints such as via-points or joint limits in the autoencoder.
To cope with this issue, we project the generated trajectory onto the constraint solution space using a trajectory optimization method.
To obtain the trajectory that satisfies the constraints, we can employ trajectory optimization methods such as CHOMP~\cite{Zucker13} and TrajOpt~\cite{Schulman14}. 
In this study, we employ the trajectory optimization method based on CHOMP.
The covariant trajectory update of CHOMP is given by
\begin{align}
\vect{\xi}^{\new} = \arg \min_{\vect{\xi}} \left\{  \mathcal{C}(\vect{\xi}^c) + \vect{g}^{\top} (\vect{\xi} - \vect{\xi}^c) + \frac{\eta}{2} \left\| \vect{\xi} - \vect{\xi}^c \right\|^2_{M}  \right\}
\label{eq:chomp}
\end{align}
where $\vect{g} = \nabla \mathcal{C}(\vect{\xi})$,$\vect{\xi}^{\new}$ is the updated plan of the trajectory, $\vect{\xi}^c$ is the current plan of the trajectory,  
$\eta$ is a regularization constant, 
and  $\left\| \vect{\xi} \right\|^2_{M}$ is the norm defined by a matrix $M$ as $\left\| \vect{\xi} \right\|^2_{M} = \vect{\xi}^{\top}M\vect{\xi}$.
The trajectory update in \eqref{eq:chomp} is equivalent to the following:
\begin{align}
\vect{\xi}^{\new} = \vect{\xi}^{c} - \frac{1}{\eta} M^{-1} \vect{g}.
\label{eq:chomp_update}
\end{align}
When the position of the end-effector at the goal position in the current plan of the trajectory deviates from the given one, we shift the goal configuration and update the entire trajectory by iterating the following update as discussed in~\cite{Dragan15}:
\begin{align}
\vect{\xi}^{\new} =  \vect{\xi}^{c} + \alpha M^{-1}[0,\ldots,0, \Delta \tilde{\vect{q}}_T]^\top,
\label{eq:projection}
\end{align}
where $\alpha$ is the learning rate and 
$\Delta \tilde{\vect{q}}_T$ is given by
\begin{align}
\Delta \tilde{\vect{q}}_T = J^{-1} 
\left[
\begin{array}{c}
\vect{x}_{\textrm{end}}(\vect{q}^0_T) - \vect{x}_{\textrm{end}}(\vect{q}^c_T) \\
\vect{0}
\end{array}
\right].
\end{align}
Constraints such as via-points and joint limits can also be incorporated in the same manner.

\begin{algorithm}[t]
	\caption{Autoencoder Trajectory Primitive (ATP) }
	\begin{algorithmic}[1]

		\STATEx{\textbf{Training phase:}}
		\STATE{\textbf{Input:} Trajectories demonstrated by experts $\{\vect{\xi}^{\demo}_i\}^M_{i=1}$ 
		}
		\STATE{Sample $N$ trajectories by following \eqref{eq:goal_dist} and \eqref{eq:exploration_stomp} }
		\STATE{Train the neural network by minimizing  $\mathcal{J}$ in \eqref{eq:loss}}
		\STATEx{\textbf{Planning phase:}}
		\STATE{\textbf{Input:} Goal position $\vect{x}_g$, and latent code $z$, $c$}
		\STATE{Generate the trajectory $\vect{\xi}^*$ with the decoder  }
		\STATE{(Optional) Project the trajectory onto the constraint solution space }
		\STATE{\textbf{Return:}  planned trajectory $\vect{\xi}^*$ }
		
	\end{algorithmic}
	\label{alg:NTP}
\end{algorithm}

\begin{figure*}[]
	\centering
	\begin{subfigure}[t]{0.465\columnwidth}
		\includegraphics[width=\textwidth]{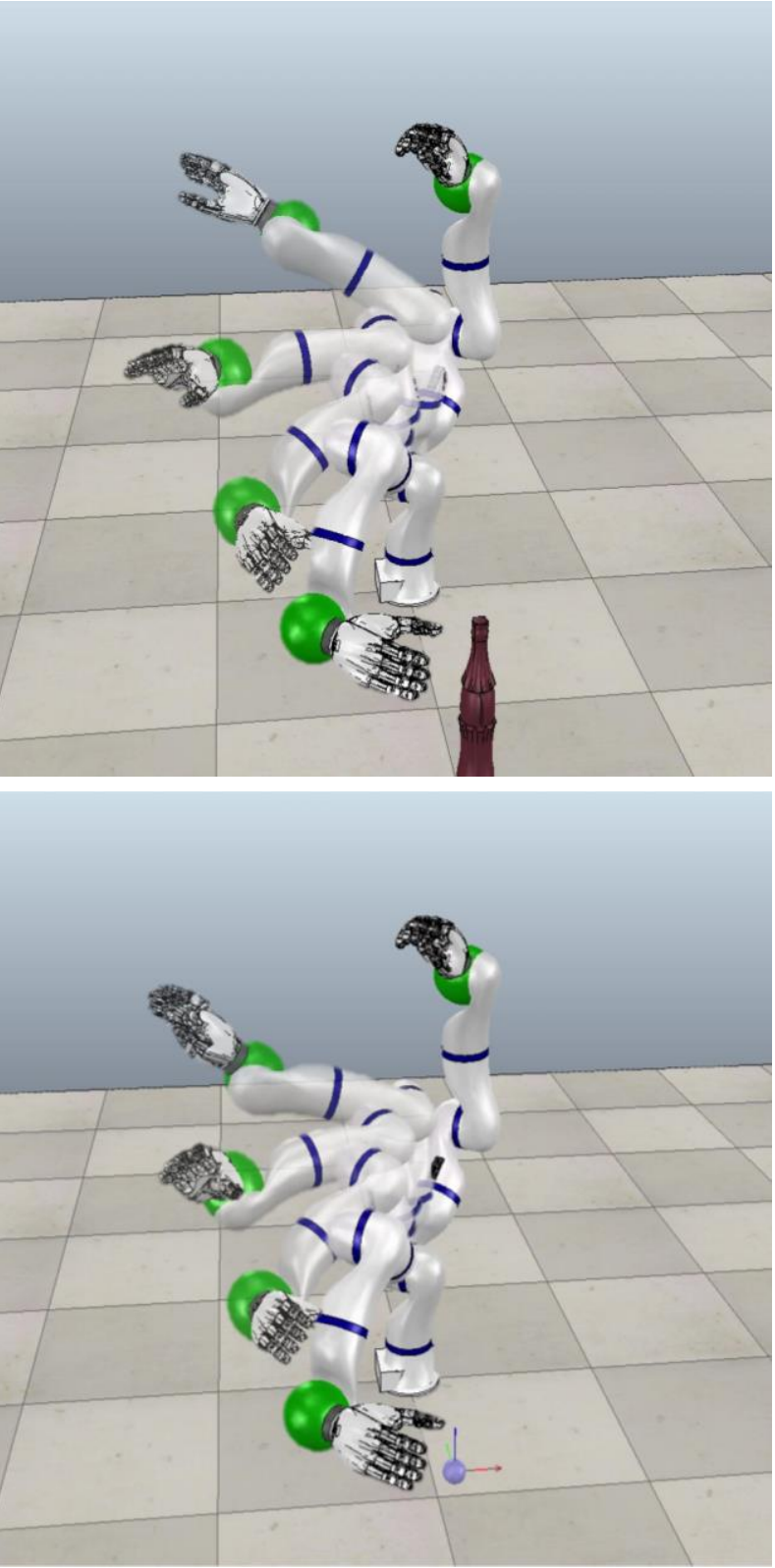}
		\caption{}
	\end{subfigure}
	\begin{subfigure}[t]{0.465\columnwidth}
		\includegraphics[width=\textwidth]{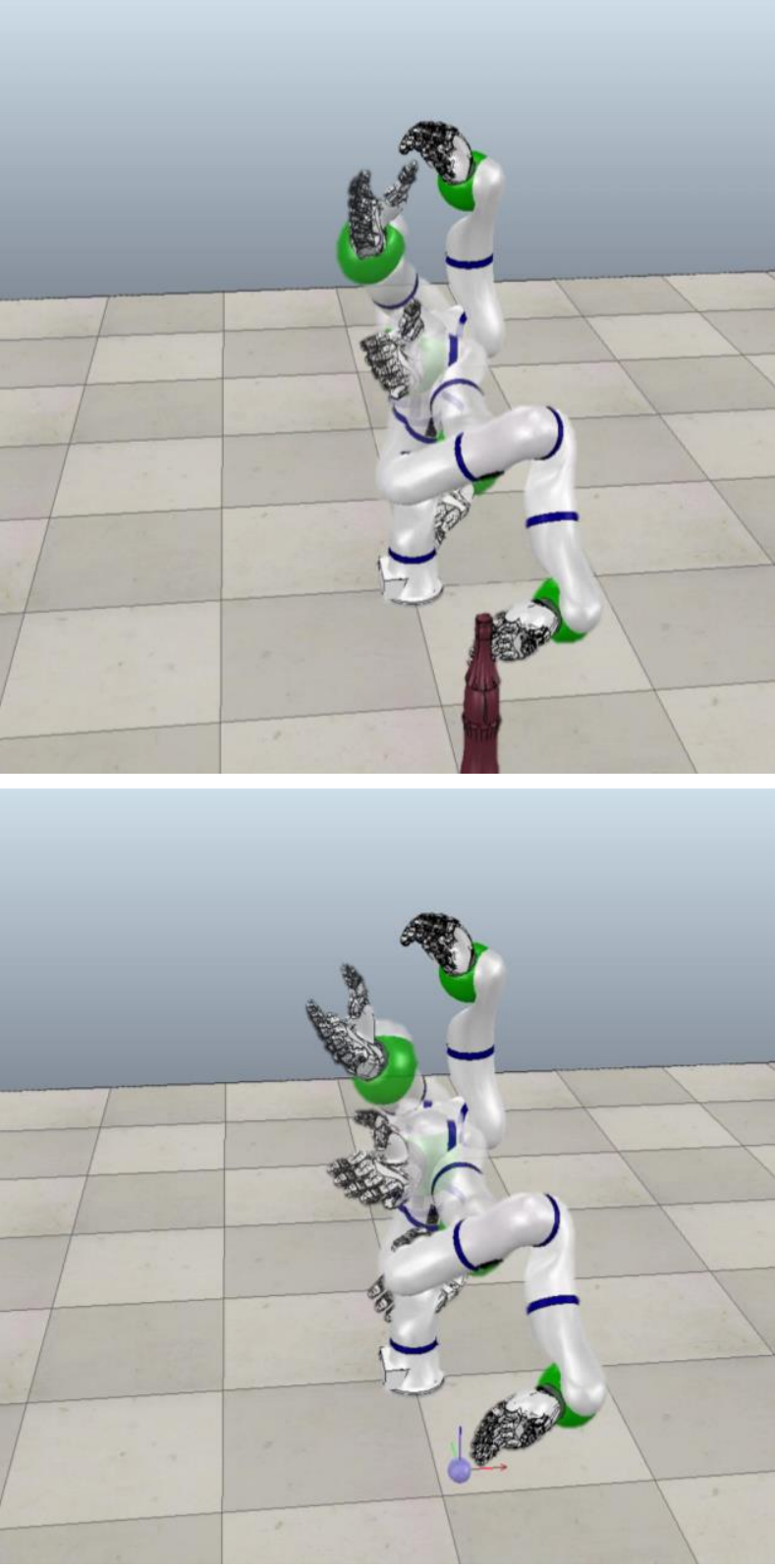}
		\caption{}
	\end{subfigure}
	\begin{subfigure}[t]{0.465\columnwidth}
		\includegraphics[width=\textwidth]{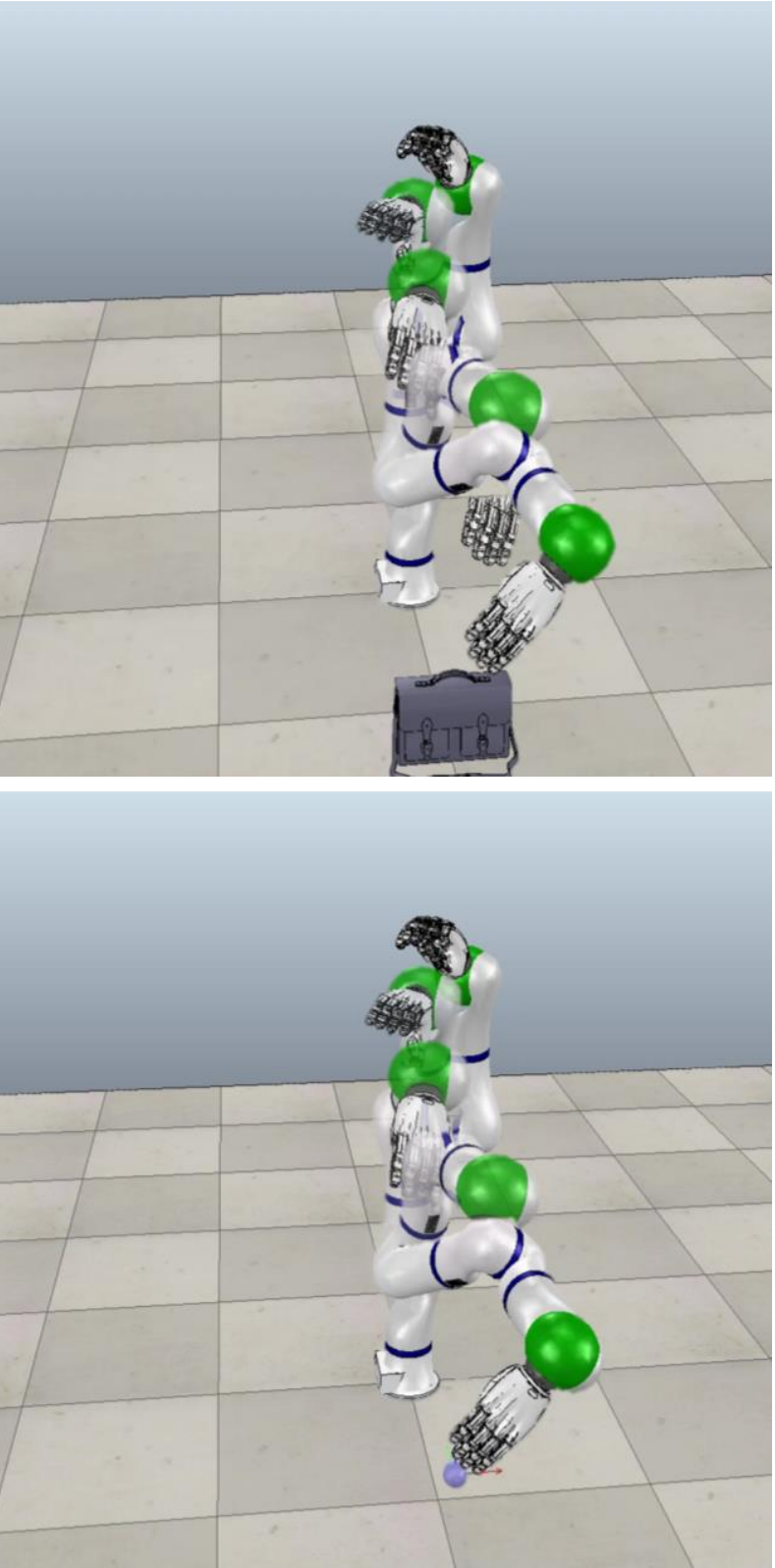}
		\caption{}
	\end{subfigure}
	\begin{subfigure}[t]{0.465\columnwidth}
		\includegraphics[width=\textwidth]{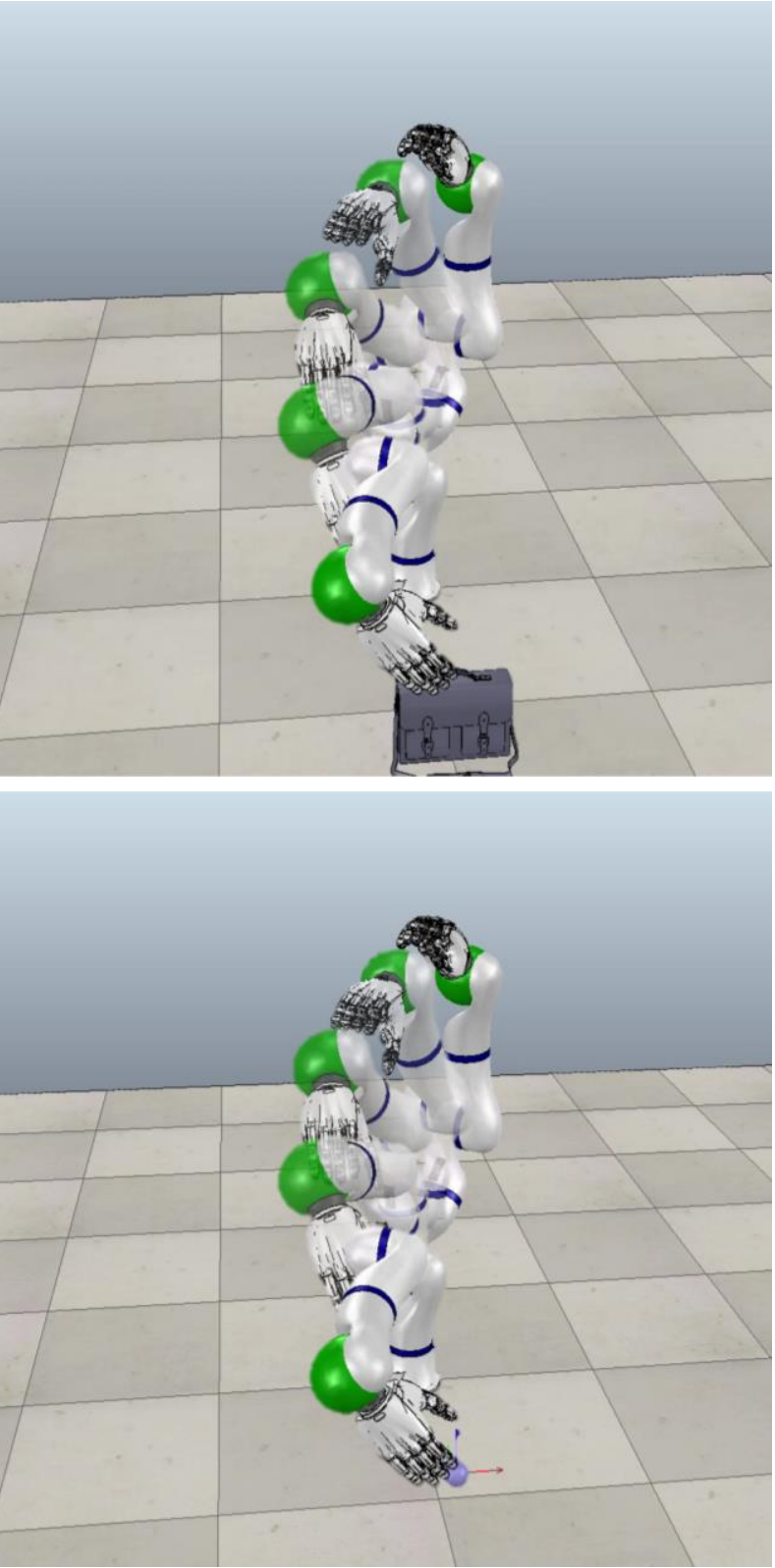}
		\caption{}
	\end{subfigure}
	\caption{Input trajectories and reconstructed trajectories. In (a)-(d), the upper figures show the input trajectories and the lower figures show the reconstructed trajectories. The blue sphere represents the goal position in task space, i.e. $\vect{x}_g = [0, -0.7, 0.2]$.
		The latent variable for reconstructing the trajectory is shown in Table~\ref{tbl:latent_reconst}.
	}
	\label{fig:reconst}
\end{figure*}

\section{Experiments}
In this experiment, four demonstrated trajectories were used.
To train the ATP, we created 4000 synthetic trajectories by using the method described in Section~\ref{sec:augmentation}.
The dimension of discrete variable $\vect{c}$ was four, and that of the continuous variable $\vect{z}$ was five.
In the following, we show the results obtained when a neural network is trained using the conditions shown in Table~\ref{tbl:cond}.
We simulated a KUKA LWR robot\footnote{We customized the color to show the motion more clearly.} with 7 degrees of freedoms.
A simulator is developed using V-REP~\cite{Rohmer13}.
To implement the variational autoencoder, we adopted the PyTorch implementation provided by the author of \cite{Dupont18}\footnote{available at https://github.com/Schlumberger/joint-vae}.

\begin{table}[b]
	\caption{Conditions for training the neural network}
	\label{tbl:cond}
	\begin{center}
		\begin{tabular}{|c|c|}
			\hline 
			Parameter & Value \\
			\hhline{|=|=|}
			\makecell{ $\#$ of epochs}  & 250 \\
			\hline
			\makecell{batch size} &  100 \\
			\hline
			\makecell{$\#$ of time step in a trajectory} &  50 \\
			\hline
			\makecell{$\#$ of units in hidden layers  in the encoder} &   (300, 200) \\
			\hline
			\makecell{Activation function in the encoder} &   (relu, relu) \\
			\hline
			\makecell{$\#$ of units in hidden layers  in the decoder} &   (200, 300) \\
			\hline
			\makecell{Activation function in the decoder} &   (relu, relu) \\
			\hline
		\end{tabular}
	\end{center}
\end{table}

Fig.~\ref{fig:reconst} shows the input trajectories and reconstructed trajectories\footnote{Please note that the frames shown in the figures are not synchronized due to the limitation of our implementation.}.
In this experiment, input trajectories are designed to represent different behavior patterns.
Figs~\ref{fig:reconst}(a) and (b) show trajectories planned for grasping a bottle with different approach angles. 
Likewise, Figs~\ref{fig:reconst}(c) and (d) show trajectories planned for grasping a bag with different orientations.
Although other frameworks often require multiple demonstrations of the same pattern, the result shows that our neural network can model multiple types of behaviors with a single model.
To  examine the quality of the reconstructed trajectories, the trajectories shown in Fig.~\ref{fig:reconst} are not projected onto the constraint solution space.
Each input trajectory is successfully reconstructed as show in Fig.~\ref{fig:reconst}. 
The latent variable generated for reconstructing the input trajectories are shown in Table~\ref{tbl:latent_reconst}.
Interestingly, the ATP encodes trajectories shown in Fig.~\ref{fig:reconst}(a) and (b) into the same discrete latent code as shown in Table~\ref{tbl:latent_reconst}.
As a result, the input trajectories shown in Fig.~\ref{fig:reconst}(a) and (b) are interpolated via the continuous latent variable $\vect{z}$ as we will see later.
The same discussion holds for the input trajectories shown in Fig.~\ref{fig:reconst}(c) and (d); these trajectories are encoded in the same discrete latent code $\vect{c} = [0,0,0,1]$, and they are interpolated via the latent variable $\vect{z}$.
Thus, the trajectories demonstrated for the same tasks are categorized into the same class represented by the discrete latent variable $\vect{c}$ in this experiment.
This result indicates that the APT can learn meaningful discrete representations in an unsupervised manner.

\begin{table}[b]
	\caption{Latent variables generated for reconstruction shown in Fig.~\ref{fig:reconst}}
	\label{tbl:latent_reconst}
	\begin{center}
		\begin{tabular}{|c|c|c|}
			\hline
			\makecell{Trajectory \\ in Fig.~\ref{fig:reconst}} & $\vect{z}$ & $\vect{c}$\\
			\hline
			(a) & [0.02, -0.25, 0.01, 0.04,-0.06] & [ 0., 0., 0., 1.] \\
			\hline
			(b) & [0.05, 0.57, -0.01,  0.04, -0.10] & [  0., 0., 0., 1.] \\
			\hline
			(c) &  [0.15, -0.30, -0.06, -0.01, -0.09] & [  1., 0., 0., 0.] \\
			\hline
			(d) &[0.12, 0.28,  -0.06, -0.03, -0.07] & [  1., 0., 0., 0.] \\
			\hline
		\end{tabular}
	\end{center}
\end{table}

\begin{figure}[]
	\centering
	\begin{subfigure}[t]{0.465\columnwidth}
		\includegraphics[width=\textwidth]{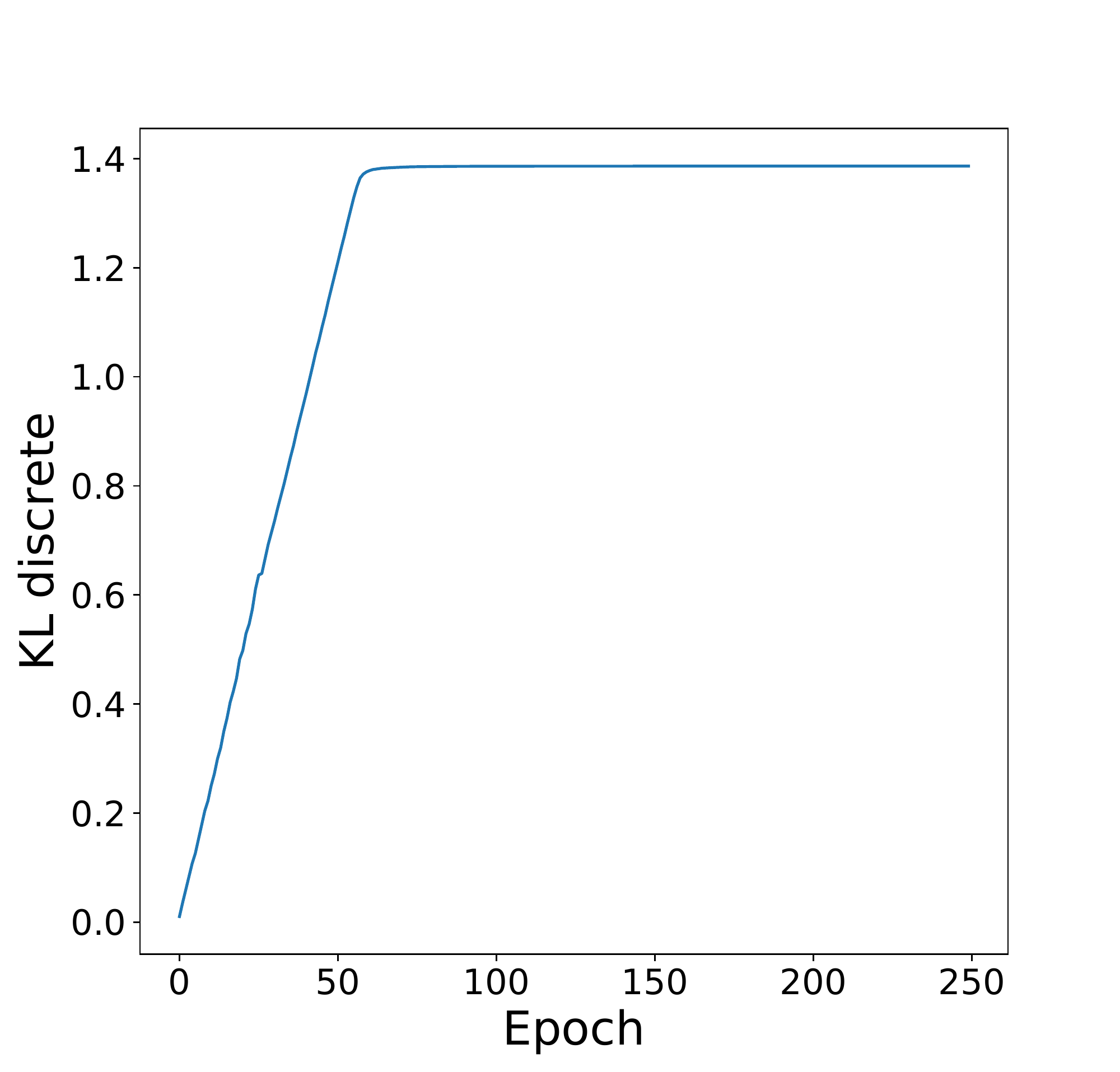}
		\caption{$\KL\big( p(\vect{c} | \vect{\xi}) || p(\vect{c}) \big)$. }
	\end{subfigure}
	\begin{subfigure}[t]{0.465\columnwidth}
		\includegraphics[width=\textwidth]{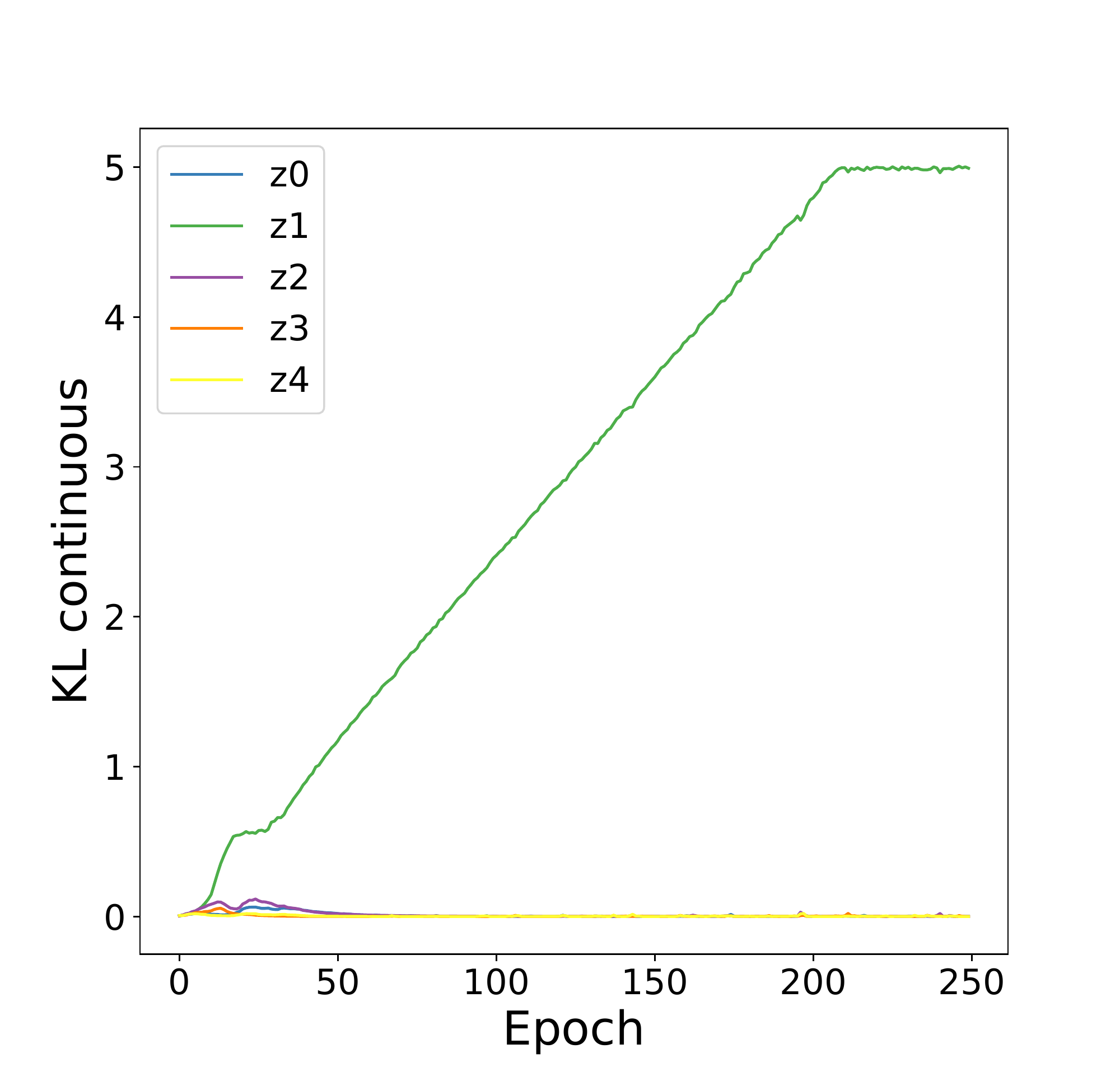}
		\caption{$\KL\big( p( z_i | \vect{\xi}) || p(z_i) \big)$ for $i=0,1,2,3,4$.}
	\end{subfigure}
	\caption{The change of KL divergence during the training of our autoencoder. }
	\label{fig:KL}
\end{figure}

As discussed in~\cite{Dupont18}, the KL divergence $\KL\big( q(\vect{z}, \vect{c} | \vect{\xi}) || p(\vect{z}, \vect{c}) \big) $ is the upper-bound of the mutual information between the latent variable and the data.
Therefore, we plot the KL divergence of each latent unit during training in Fig.~\ref{fig:KL} as in~\cite{Dupont18}.
As shown, among the five continuous variables, only one channel $z_1$ has a significant value; other four channels do not.
This result indicates that we can obtain various trajectories that cover the training dataset by selecting one of four discrete variable and changing only $z_1$. 

\begin{figure}[]
	\centering
	\includegraphics[width=0.8\columnwidth]{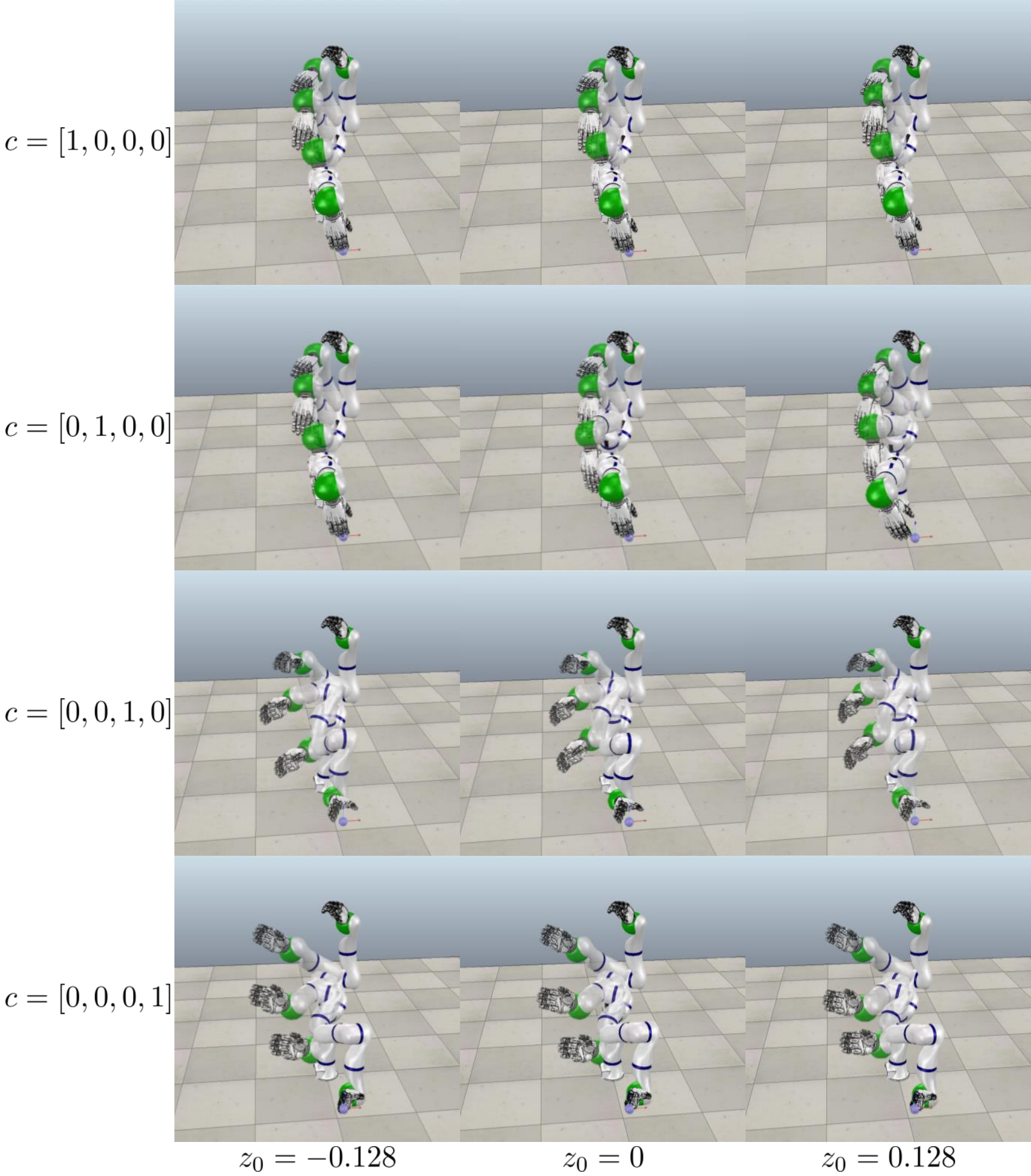}
	\caption{Trajectories generated by changing $z_0$. From the top, each row corresponds to $\vect{c}=[1, 0, 0, 0]$, $\vect{c}=[0, 1, 0, 0]$, $\vect{c}=[0, 0, 1, 0]$, $\vect{c}=[0, 0, 0, 1]$, respectively. $z_1 = z_2 = z_3 = 0$ and $\vect{x}_g = [0, -0.7, 0.2]$ in all trajectories. The column corresponds to varying $z_0$.  }
	\label{fig:traverse_z0}
\end{figure}

The trajectories obtained by changing $\vect{c}$ and $z_0$ are shown in Fig.~\ref{fig:traverse_z0}.
It is evident that the change of $z_0$ does not result in a significant change of the generated trajectories.
This result is reasonable because the log of the KL divergence in Fig.~\ref{fig:KL} indicates that $z_0$ does not contain much information.

\begin{figure*}[]
	\centering
	\includegraphics[width=\textwidth]{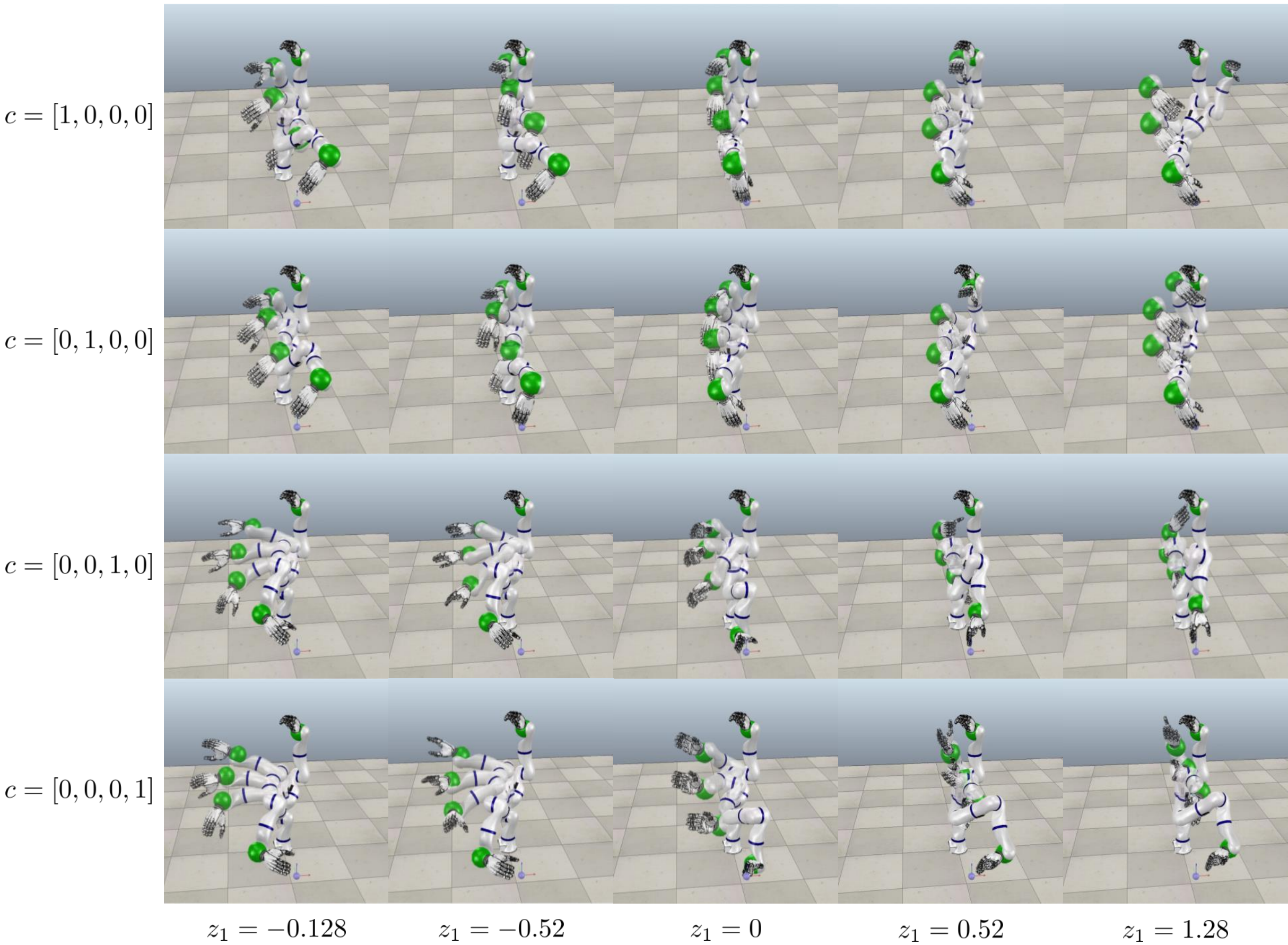}
	\caption{Trajectories generated by changing $z_1$. From the top, each row corresponds to $\vect{c}=[1, 0, 0, 0]$, $\vect{c}=[0, 1, 0, 0]$, $\vect{c}=[0, 0, 1, 0]$, $\vect{c}=[0, 0, 0, 1]$, respectively. $z_0 = z_2 = z_3 = 0$  and $\vect{x}_g = [0, -0.7, 0.2]$ in all trajectories. The column corresponds to varying $z_1$. }
	\label{fig:traverse_z1}
\end{figure*}

The trajectories obtained by changing $\vect{c}$ and $z_1$ are shown in Fig.~\ref{fig:traverse_z1}.
It is clear that the change of $z_1$ results in significant change of the shapes of generated trajectories.
This result also fits with the result that only $z_1$ contains significant information as shown in Fig.~\ref{fig:KL}.
In the top row of Fig.~\ref{fig:traverse_z1}, which corresponds to $\vect{c} =[1,0,0,0] $, one can see that the input trajectories shown in Fig.~\ref{fig:reconst}(a) and (b) are continuously interpolated via the latent variable $\vect{z}$.
Accordingly, the trajectory shown in the middle of the top row of Fig.~\ref{fig:traverse_z1} appears to be a mixture of the trajectories shown in Fig.~\ref{fig:reconst}(a) and (b).
Likewise, it can be seen in the bottom row of Fig.~\ref{fig:traverse_z1}, which corresponds to $\vect{c} =[0,0,0,1] $, that the input trajectories shown in Fig.~\ref{fig:reconst}(c) and (d) are continuously interpolated via the latent variable $\vect{z}$.
As a result, the middle of the bottom row of Fig.~\ref{fig:traverse_z1} appears to be a mixture of the trajectories shown in Fig.~\ref{fig:reconst}(c) and (d).

\begin{figure}[]
	\centering
	\includegraphics[width=\columnwidth]{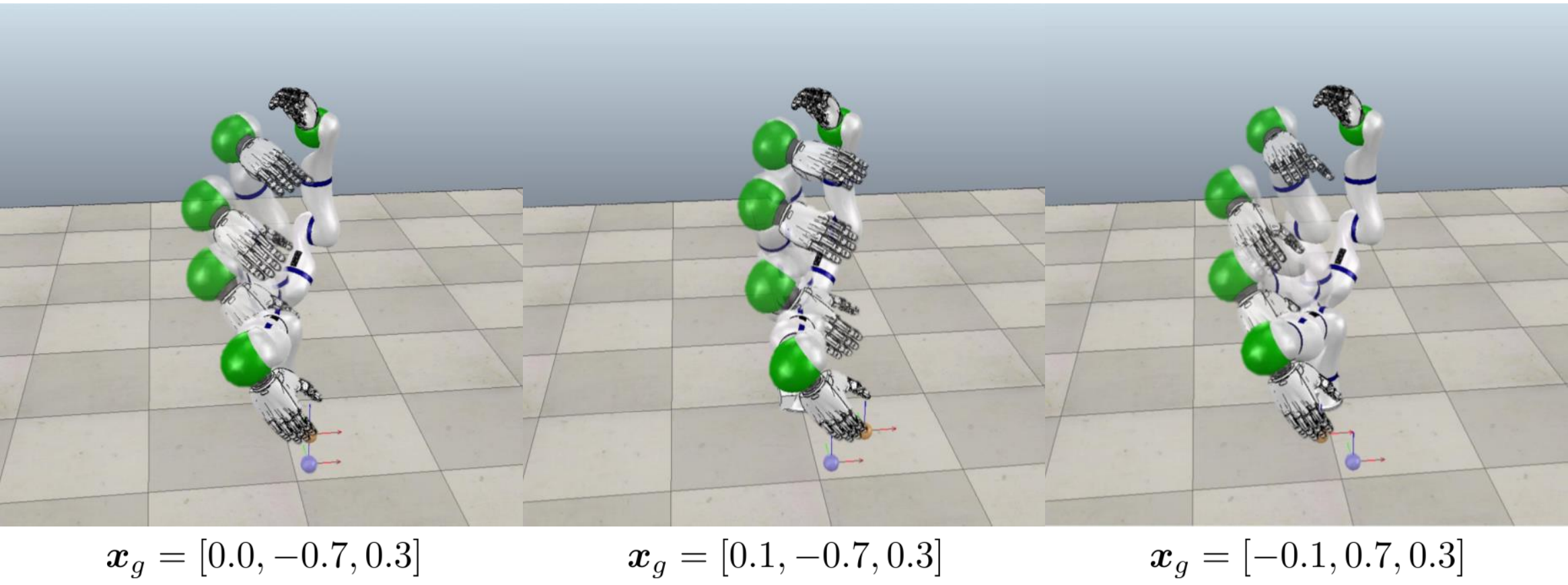}
	\caption{Trajectories generated by changing $\vect{x}_g$. The trajectories are properly generalized to different goal positions.
		$\vect{c}=[1,0,0,0]$ and $\vect{z} = [0., 1.28, 0., 0., 0.]$ in all trajectories. The orange sphere indicates the specified goal position, and the blue sphere indicates the goal position used in the demonstrated trajectories. The unit of $\vect{x}_g$ is the meter in this simulation.}
	\label{fig:varying_goal}
\end{figure}
The trajectories obtained by changing $\vect{x}_g$ are shown in Fig.~\ref{fig:varying_goal}.
The trajectories are properly generalized to different goal positions of the end-effector.
As discussed in \cite{Cremer18}, the output of the variational autoencoder is amortized variational inference.
Therefore, the generalization with the decoder is not accurate, e.g. the positioning error is approximately from 0.01 to 0.05~m.
However, after the projection with CHOMP, the positioning error at the goal position can be significantly small, e.g., less than 1mm.
This results show that the ATP can generalize the learned trajectories to different goal positions.

From these results, it is evident that our autoencoder learned interpretable latent representations from a limited number of demonstrated trajectories.
On the other hand, the hand orientation and the shape of the entire trajectory is entangled in the learned latent variable;
when changing $z_1$, both the orientation and the shape of the entire trajectory change.
In practice, a user may need to tune the shape of the entire trajectory without changing the hand orientation, e.g. for collision avoidance.
However, when using the latent variable learned in this study, it is not possible to change the shape of the entire trajectory without changing the hand orientation.
We think that this result is due to the property of the objective function of our neural network and that this point needs to be addressed in future work.

\section{Discussion}
The results indicate that the ATP learns useful latent representations for tuning the trajectory shape. 
The learned decoder can be used as a motion planner that takes the goal position and latent variable as its input.
Using the ATP, a user can teach the desired motion by performing a handful of demonstrations, and when a trajectory is planned for a new task, a user can also select and modify the trajectory by changing the discrete and continuous latent variables.
The trajectory augmentation presented in this paper can be viewed as a way to autonomously explore the trajectory for learning useful low-dimensional representations.

Existing movement primitive frameworks, such as DMP and ProMP, often require separate models for representing diverse behaviors.
In other words, these frameworks require label information for learning multiple types of behaviors.
As a result, if a given demonstration dataset contains multiple types of behaviors without labels, they often fails to model the given trajectories.
However, as shown by the experimental results, the ATP can model diverse behavior by encoding the trajectory types into the latent variables in an unsupervised manner.
This property of the ATP is beneficial in practice, since it will reduce the effort to consider the types of demonstrated trajectories.
Prior work such as~\cite{Paraschos13} considered the combination of movement primitives by introducing the activation factor.
In contrast, ATP can deal with the blending of multiple types of behaviors through the continuous latent variable, which does not require learning multiple models.
This property is also useful in practice when the user need to plan a trajectory which is similar to input trajectories but different from them.

Neural networks are often considered to be incomprehensible due to their complexity. 
However, our work shows that neural networks can provide a way to interpret and manipulate incomprehensible high-dimensional data by finding low-dimensional latent representations.
Although trajectory planning in robotic systems often require expert knowledge, the low-dimensional representation found by our framework can be used to tune the trajectory shape without expert knowledge.

We think that the proposed framework can be used for task-level motion planning.
The task-level motion planning is often built on pre-fixed motion planner, and it is not trivial to tune each lower-level planner.
However, by using our framework, lower-level trajectories can be represented by low-dimensional information and the tuning of such a low-dimensional vector should be much easier compared to the raw representations of trajectories.

\section{Conclusions}
We proposed the autoencoder trajectory primitive~(ATP), which is a framework for modeling demonstrated trajectories with a neural network.
In the proposed framework, the latent variables that encodes the multiple types of trajectories are learned in an unsupervised manner.
The trajectory augmentation trick was proposed to address the issue of the size of the training data.
The learned decoder can be used as a motion planner in which the user can specify the goal position and the trajectory types by setting the latent variables. 
Our experimental results show that a neural network can be trained with a handful of demonstrated trajectories and that the ATP successfully learns discrete and continuous low-dimensional latent variables.





\section*{ACKNOWLEDGMENT}

This work is supported by KAKENHI 19K20370 and 18H01410.


\bibliographystyle{IEEEtran}
\bibliography{TrajAE}

\end{document}